\documentclass[12pt]{article}
\usepackage[bookmarks=true]{hyperref}
\usepackage{times}
\usepackage{graphicx} % For figures
\usepackage{xcolor} % For Red Text
\usepackage{placeins} % For FloatBarrier
\usepackage{multirow} % For multirow table
\usepackage{caption} % For suppressing captions (and other settings)

\usepackage{setspace} % For double-space
\usepackage{hyperref} % For link

\graphicspath{{Figures/}} % Default Graphics path
\usepackage{lineno}
\usepackage[export]{adjustbox}
\usepackage{soul,xcolor}
\setstcolor{red}

\title{A soft robot that adapts to environments through shape change}

\author{Dylan S. Shah$^{1}\dagger$,  Joshua P. Powers$^{2}\dagger$, \\
Liana G. Tilton$^{1}$, Sam Kriegman$^{2}$, Josh Bongard$^{2}$, and Rebecca Kramer-Bottiglio$^{1\ast}$
\\
$^{1}$Department of Mechanical Engineering \& Materials Science, Yale University.
\\
$^{2}$Department of Computer Science, University of Vermont.
\\
$\dagger$These authors contributed equally.
\\
*Corresponding Author: {\tt rebecca.kramer@yale.edu}.
}

\makeatletter
\newcommand*\titleheader[1]{\gdef\@titleheader{#1}}
\AtBeginDocument{%
  \let\st@red@title\@title
  \def\@title{%
    \bgroup\normalfont\large\centering\@titleheader\par\egroup
    \vskip1.5em\st@red@title}
}
\makeatother

\titleheader{The definitive version of this article was published in Nature Machine Intelligence (2020). URL: nature.com/articles/s42256-020-00263-1}

\date{}
\begin{document}

\maketitle

%%%%%%%%%% Abstract %%%%%%%%%%
\begin{abstract}
Many organisms, including various species of spiders and caterpillars, change their shape to switch gaits and adapt to different environments. Recent technological advances, ranging from stretchable circuits to highly deformable soft robots, have begun to make shape-changing robots a possibility. However, it is currently unclear how and when shape change should occur, and what capabilities could be gained, leading to a wide range of unsolved design and control problems. To begin addressing these questions, here we simulate, design, and build a soft robot that utilizes shape change to achieve locomotion over both a flat and inclined surface. Modeling this robot in simulation, we explore its capabilities in two environments and demonstrate the existence of environment-specific shapes and gaits that successfully transfer to the physical hardware. We found that the shape-changing robot traverses these environments better than an equivalent but non-morphing robot, in simulation and reality.
\end{abstract}

%%%%%%%%%% Introduction %%%%%%%%%%
\section{Introduction}
Nature provides several examples of organisms that utilize shape change as a means of operating in challenging, dynamic environments.
For example, the spider Araneus Rechenbergi~\cite{jager_cebrennus_2014,bhanoo_desert_2014} and the caterpillar of the Mother-of-Pearl Moth (Pleurotya ruralis)~\cite{armour_rolling_2006} transition from walking gaits to rolling in an attempt to escape predation. Across larger time scales, caterpillar-to-butterfly metamorphosis enables land to air transitions, while mobile to sessile metamorphosis, as observed in sea squirts, is accompanied by radical morphological change. Inspired by such change, engineers have created caterpillar-like rolling~\cite{lin_goqbot:_2011}, modular~\cite{christensen2006evolution, yim2007modular, parrott2018hymod}, tensegrity~\cite{paul2006design, sabelhaus2015system}, plant-like growing~\cite{sadeghi_toward_2017}, and origami~\cite{miyashita2015untethered, rus2018design} robots that are capable of some degree of shape change. However, progress toward robots which dynamically adapt their resting shape to attain different modes of locomotion is still limited. Further, design of such robots and their controllers is still a manually intensive process.

% Machines in multiple environments
Despite the growing recognition of the importance of morphology and embodiment on enabling intelligent behavior in robots~\cite{pfeifer_self-organization_2007}, most previous studies have approached the challenge of operating in multiple environments primarily through the design of appropriate control strategies. For example, engineers have created robots which can adapt their gaits to locomote over different types of terrain~\cite{saranli2001rhex, raibert2008bigdog, kuindersma2016optimization}, transition from water to land ~\cite{ijspeert2007swimming, li2015design}, and transition from air to ground~\cite{myeong2015development, bachmann2009biologically, roderick2017touchdown}. Other research has considered how control policies should change in response to changing loading conditions~\cite{korayem2010dynamic, li2015discrete}, or where the robot's body was damaged~\cite{bongard2006resilient, cully2015robots, chatzilygeroudis2018reset}. Algorithms have also been proposed to exploit gait changes that result from changing the relative location of modules and actuators~\cite{rosendo_trade-off_2017}, or tuning mechanical parameters, such as stiffness~\cite{garrad_shaping_2018}. In such approaches, the resting dimensions of the robot's components remained constant. These robots could not, for instance, actively switch their body shape between a quadrupedal form and a rolling-optimized shape.

% Soft robotics
The emerging field of soft robotics holds promise for building shape-changing machines~\cite{hauser_resilient_2019}. For example, one robot switched between spherical and cylindrical shapes using an external magnetic field, which could potentially be useful for navigating internal organs such as the esophagus and stomach~\cite{yim2012shape}. Robotic skins wrapped around sculptable materials were shown to morph between radially-symmetric shapes such as cylinders and dumbbells to use shape-change as a way to avoid obstacles~\cite{shah2019morphing}. Lee et al. proposed a hybrid soft-hard robot that could enlarge its wheels and climb onto step-like platforms~\cite{lee_origami_2017}. A simulated soft robot was evolved to automatically regain locomotion capability after unanticipated damage, by deforming the shape of its remnant structure~\cite{kriegman2019automated}. With the exception of the study by Kriegman et al.~\cite{kriegman2019automated}, control strategies and metamorphosis were manually programmed into the robots, thereby limiting such robots to shapes and controllers that human intuition is capable of designing. However, there may exist non-intuitive shape-behavior pairings that yield improved task performance in a given environment. Furthermore, manufacturing physical robots is time-consuming and expensive relative to robot simulators such as VoxCad~\cite{hiller2014dynamic}, yet discovering viable shape-behavior pairs and transferring simulated robots to functioning physical hardware remains a challenge. Although many simulation to reality (``sim2real'') methods have been reported~\cite{jakobi1995noise, lipson2000automatic, bongard2006resilient, koos2013transferability, cully2015robots, bartlett20153d, rusu_sim--real_2017, chebotar_closing_2019, peng2018sim,  hwangbo2018learning, hiller2012automatic},
none have documented the transfer from simulation to reality of shape-changing robots.

\begin{figure}
    \centering
    \includegraphics[width=5 in]{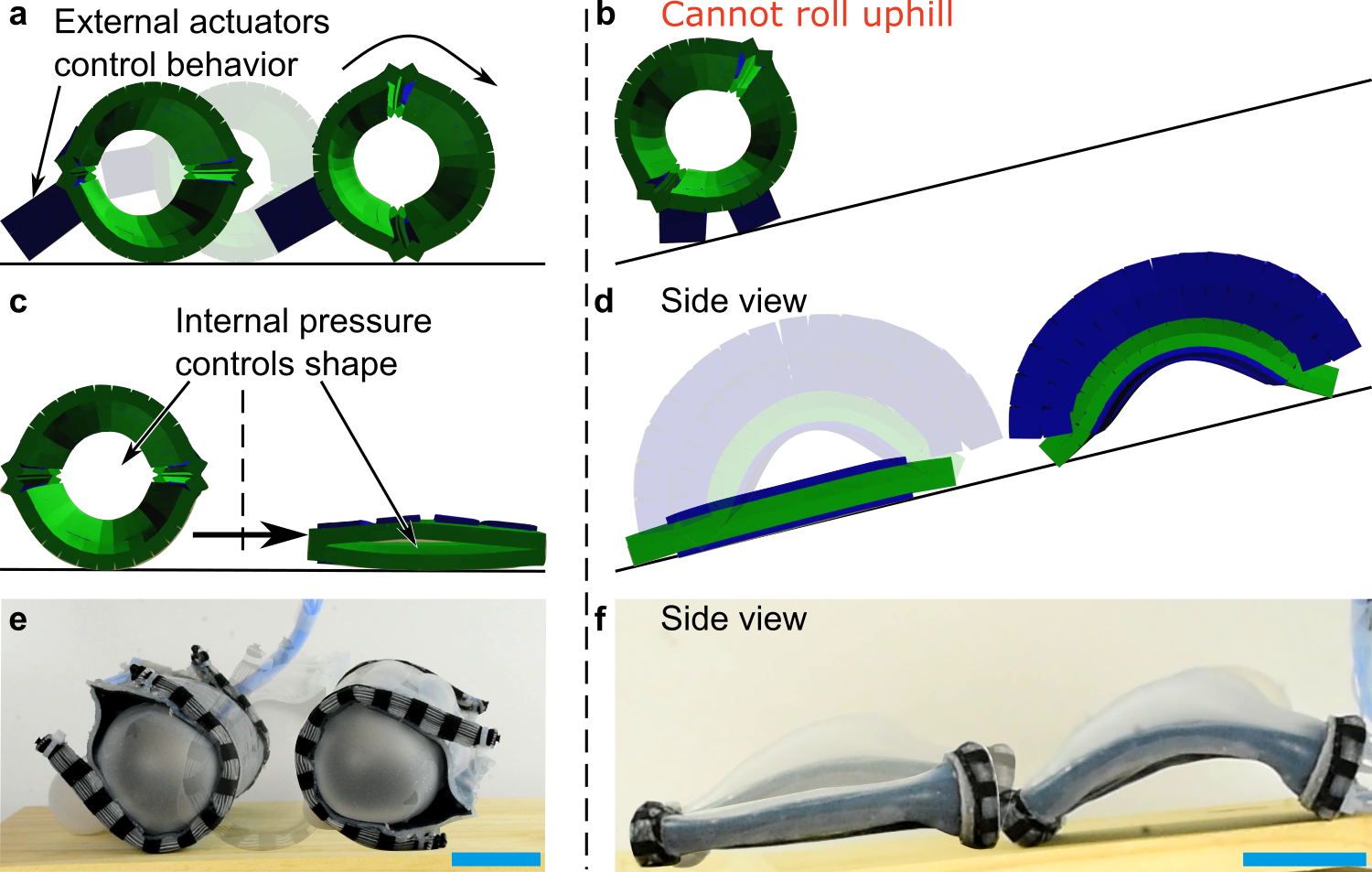}
    \caption{\textbf{Shape change can result in faster locomotion speeds than control adaptation, when a robot must operate in multiple environments.} \textbf{a,} Using inflatable external bladders, rolling was the most effective gait on flat ground. \textbf{b,} Rolling was ineffective on the inclined surface.
    We discovered
    a flat shape (achieved by deflating the inner bladder; \textbf{c}) and crawling gait (\textbf{d}) that allowed the robot to succeed in this environment. \textbf{e,f,} After
    we discovered
    these strategies in simulation, we transferred
    these
    strategies for rolling (\textbf{e}) and inchworm motion (\textbf{f}) to real hardware. Scale bars, 5 cm.}
    \label{fig: summary}
\end{figure}

\begin{figure}
    \centering
    \includegraphics[width=5 in]{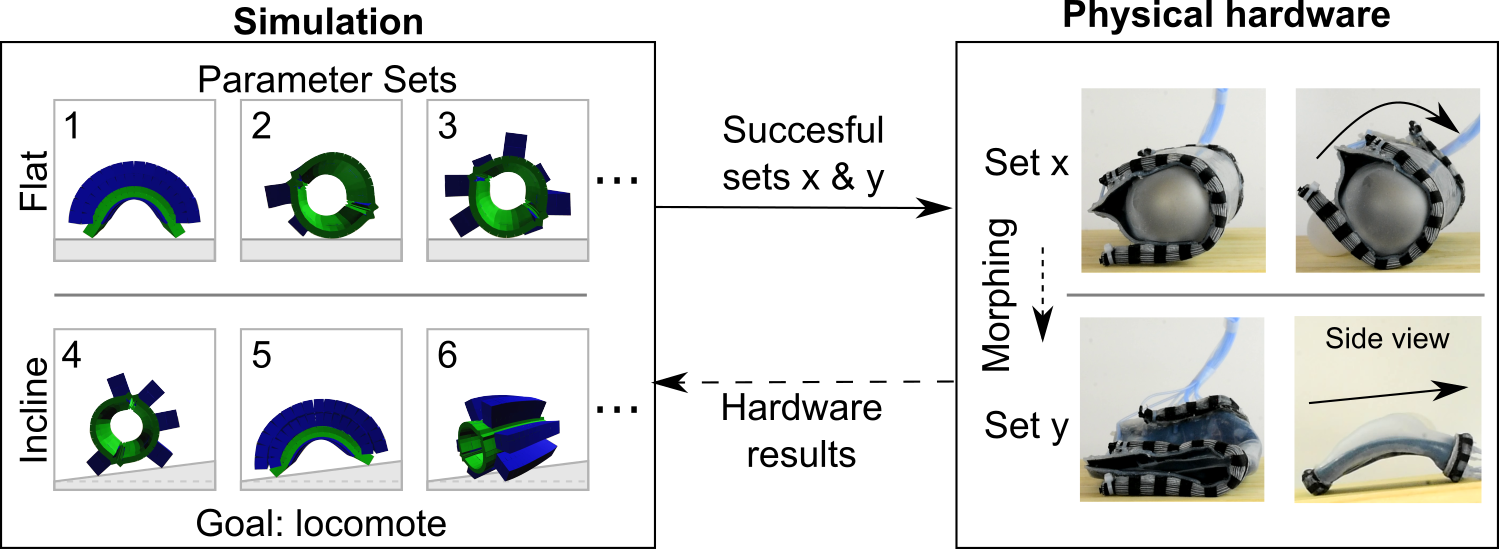}
    \caption{\textbf{Simulation revealed successful shapes and controllers, which we attempted to realize in hardware.} Sets consisting of a shape, an orientation, and a controller were generated for the robot in simulation. Each numbered sub-panel depicts a single automatically generated parameter set. After running simulations to determine the speed of each set, some were deemed too slow, while successful (relatively quick) sets were used to design a single physical robot that could reproduce the shapes and gaits found in simulation for both environments. During prototyping, actuator limits were measured and incorporated into the simulator to improve the accuracy of the simulation.}
    \label{fig: sim2real}
\end{figure}

% describe problem set-up more and what we are trying solve
To test whether situations exist where shape change improves a robot's overall average locomotion speed within a set of environments more effectively than control adaptations, here we present a robot which actively controls its shape to locomote in two different environments: flat and inclined surfaces (Fig.~\ref{fig: summary}). The robot had an internal bladder, which it could inflate/deflate to change shape, and a single set of external inflatable bladders which could be used for locomotion. Depending on the core's shape, the actuators created different motions, which could allow the robot to develop new gaits and gain access to additional environments. Within a soft multi-material simulator, an iterative ``hill-climbing'' algorithm~\cite{mitchell_when_1994} generated multiple shapes and controllers for the robot, then automatically modified the robots' shapes and controllers to discover new locomotion strategies. No shape-controller pairs were found that could locomote efficiently in both environments. However, even relatively small changes in shape could be paired with control policy adaptations to achieve locomotion within the two environments. In flat and even slightly inclined environments, the robot's fastest strategy was to inflate and roll. At slopes above a critical transition angle,
we could increase the robot’s speed by flattening it and programming it with an inchworm gait
to exhibit an inchworm gait. A physical robot was then designed and manufactured to achieve similar shape-changing ability and gaits (Fig.~\ref{fig: sim2real}). When placed in real-world analogs of the two simulated environments, the physical robot was able to change shape to locomote with two distinct environmentally-effective gaits, demonstrating that shape change is a physically realistic adaptation strategy for robots.

% Sim & real summary

% Sim2real loop

This work points toward the creation of a pipeline that takes as input a desired objective within specified environments, automatically searches in simulation for appropriate shape and control policy pairs for each environment, and then searches for transformations between the most successful shapes. If transformations between successful shapes can be be found, those shape/behavior pairs are output as instructions for designing the metamorphosing physical machine. Here, we demonstrate that at least some shape/behavior pairs, as well as changes between shapes, can be transferred to reality. Thus, this work represents an important step toward an end-to-end pipeline for shape-changing soft robots that meet the demands of dynamic, real-world environments.

%%%%%%%%%% The simulated robot. %%%%%%%%%%
\section{Results}
\subsection{The simulated robot.}
\begin{figure} % Simulation results in graph format
    \centering
    \includegraphics[width=\textwidth]{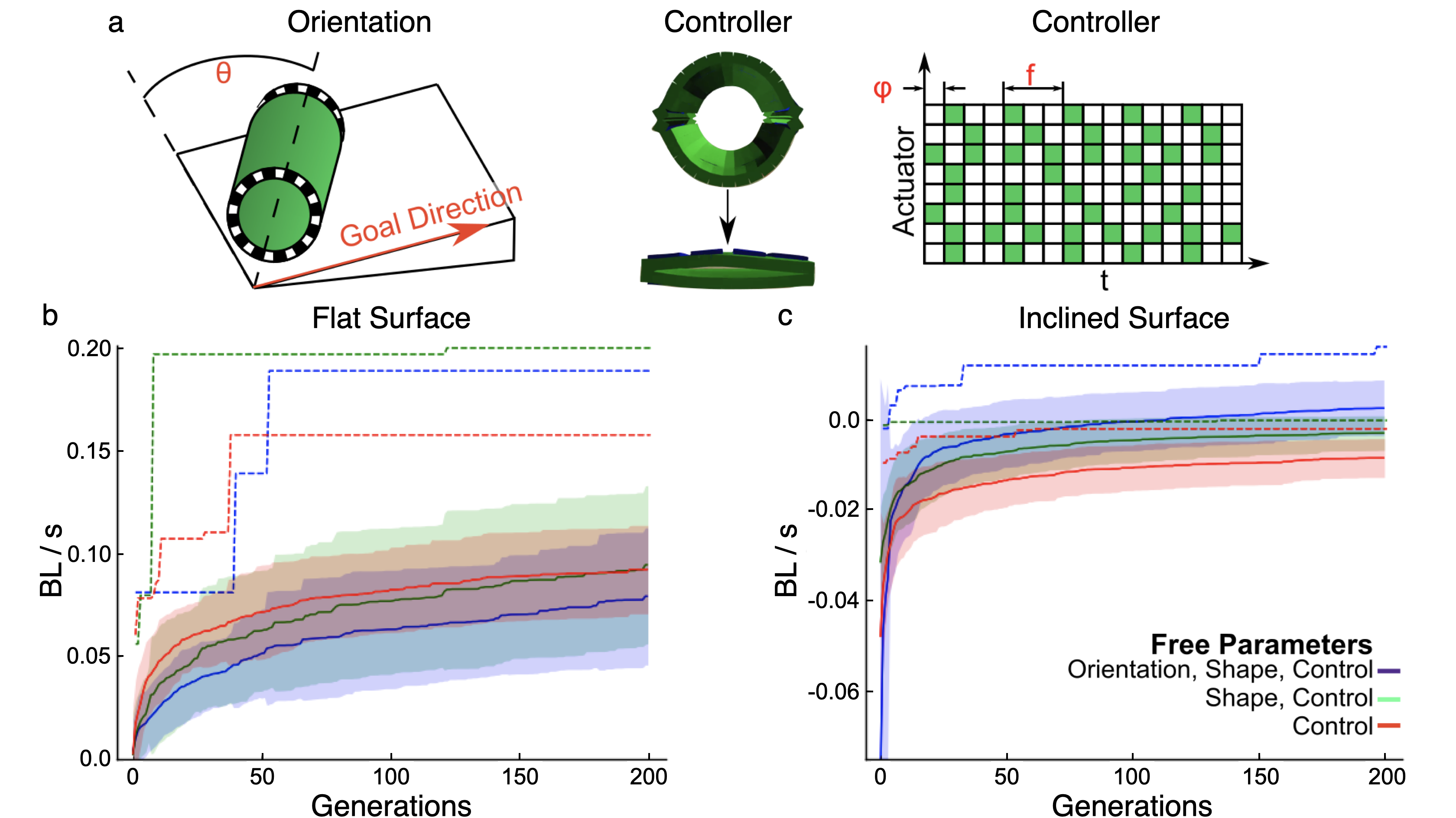}
    \caption{
    \textbf{Automated search attempting to improve gaits in both environments.}
    \textbf{a,} For each simulation, the algorithm could adjust the orientation, shape, and/or controller of the robot. Orientation ($\theta$) was measured by the angle between the robot's leading edge and a constant-elevation line on the surface. Shape was parameterized as the inner bladder's pressure, resulting in a family of shapes between the cylinder and flat shape shown. Control of each actuator was parameterized as the number of timesteps until its first actuation ($\phi$) and the number of timesteps between actuations ($f$). Here, we show an example controller for the eight main bladders, with green shaded squares illustrating inflation and white squares showing deflation.
    \textbf{b,} Results on a flat surface and \textbf{(c)} on an inclined surface. Shaded regions represent one standard deviation about the mean (solid line), while dashed lines represent maximum fitness. The legend indicates which parameters were to open to optimization, the others being held constant.
    }
    \label{fig:sim_results}
\end{figure}

% Intro to simulation
We sought
to automate search for efficient robot shapes and control policies in simulation, to test our hypothesis that shape and controller adaptation can improve locomotion speeds across changing environments more effectively when given a fixed amount of computational resources, as compared to controller adaption only. To verify that multiple locomotion gaits were possible with the proposed robot design, we first used our intuition to create two hand-designed shape and control policies: one for rolling while inflated in a cylindrical shape (Fig.~\ref{fig: summary}a), and the other for inchworm motion while flattened (Fig.~\ref{fig: summary}d). Briefly, the rolling gait consisted of inflating the trailing-edge bladder to tip the robot forward, then inflating one actuator at a time in sequence. The hand-designed ``inchworm'' gait consisted of inflating the four upward-facing bladders simultaneously to bend the robot in an arc. We then performed three pairs of experiments in simulation. Within each pair, the first experiment automatically sought robot parameters for flat ground; the second experiment sought parameters for the inclined plane. Each successive pair of experiments allowed the optimization routine to control an additional set of the robot's parameters, allowing us to measure the marginal benefit of adapting each parameter set when given identical computational resources (summarized in Table~\ref{tbl: simulation results} and Fig.~\ref{fig:sim_results}). The three free parameter sets of our shape-changing robot are shape, orientation relative to the contour (equal-elevation) lines of the environment, and control policy (Fig.~\ref{fig:sim_results}a). This sequence of experiments sought to determine whether optimization could find successful parameter sets in a high-dimensional search space, while also attempting to determine to what degree shape change was necessary and beneficial.

% Fitness
In all experiments, fitness was computed as the distance the simulated robot traveled in a desired direction for a fixed time period. To facilitate comparison with speeds achieved by the physical robot, the average speed of the simulated robot, when using hand-designed or evolved shapes and policies, was computed and is reported herein in body lengths/second (BL/s) attained over flat ground or uphill, depending on the current environment of interest, during a fixed period of time. 

Parameters for the simulation were initialized based on observations of previous robots~\cite{shah2019morphing,booth_omniskins:_2018}, and adjusted to reduce the simulation-to-reality gap after preliminary tests with physical hardware (see Methods for additional details). The results reported here are for the final simulations that led to the functional physically-realized robot and gaits.

\begin{table}[ht] % Multicolumn table https://tex.stackexchange.com/questions/188690/how-to-make-a-table-with-subcolumns-like-this
\small
  \caption{\textbf{Simulation results, reported as the mean and maximum velocity attained for each test condition. The simulator is deterministic, so no mean is reported for the hand-designed gaits (since they will always yield identical locomotion speed). Shape-change allowed the robot to switch between dissimilar locomotion gaits, outperforming the benchmark policies. Combined maximum was determined by averaging the maximum speed attainable in both environments. All values have units of body-lengths per second (BL/s).}}
  \label{tbl: simulation results}
  \begin{centering}
  \begin{tabular}{|p{5cm}|c|c|c|c|c|}
    \hline
    \multirow{2}{5cm}{Free parameters} & \multicolumn{2}{c|}{\textbf{Flat ground}} & \multicolumn{2}{c|}{\textbf{Hill}} & ~\\
    \cline{2-5}
    & Mean & Max & Mean & Max & Combined max\\
    \hline
    %Orientation, Shape, Control & 0.0790730346450634 & 0.188832647888239 & 0.00226589568399288 & 0.0158172698852103 & 0.1023249588867245\\
    Orientation, Shape, Control & 0.0791 & 0.1888 & 0.0023 & 0.0158 & 0.1023\\
    \hline
    %Shape, Control & 0.0945451890675152 & 0.199979314420894 & -0.00320097179706825 & -0.000426222040531251 & 0.0997765461901815\\
    Shape, Control & 0.0945 & 0.2000 & -0.0032 & -0.0004 & 0.0998\\
    \hline
    %Control & 0.0921274988912492 & 0.157649233065188 & -0.00870513768964238 & -0.00227232601960259 & 0.0776884535227925\\
    Control & 0.0921 & 0.1576 & -0.0087 & -0.0023 & 0.0777\\
    \hline
    %Hand-designed rolling & N/A & 0.175313607559892 & N/A & -0.574308147077608 & -0.199497269758858\\
    Hand-designed rolling & N/A & 0.1753 & N/A & -0.5743 & -0.1995\\
    \hline
    %Hand-designed inching & N/A & 0.0602816654437786 & N/A & 0.024556217516104 & 0.0424189414799413\\
    Hand-designed inching & N/A & 0.0603 & N/A & 0.0246 & 0.0424\\
    \hline
  \end{tabular}
  \end{centering}
\end{table}

% Summary of experiments
In the first pair of experiments, we sought to discover whether optimization could find any viable controllers within a constrained optimization space, which was known to contain the viable hand-designed controllers. Solving this initial challenge served to test the pipeline prior to attempting to search in the full search space, which has the potential to have more local minima. The shape and orientation were fixed (flat and oriented length-wise, $\theta=90{^\circ}$, for the inclined surface, cylindrical and oriented width-wise, $\theta=0{^\circ}$ for the flat surface). In the second pair of experiments, the algorithm was allowed to simultaneously search for an optimal shape and controller pair. Finally, in the third pair of experiments, all three parameter sets were open to optimization in both environments, allowing optimization the maximum freedom to produce novel shapes, orientations, and controllers for locomoting in the two different environments.
For each experiment, we ran 60 independent ``hill-climbers'' (instantiations  of the ``hill-climbing'' search algorithm~\cite{mitchell_when_1994}, not to be confused with a robot that climbs a hill) for 200 generations, thus resulting in identical resource allocation for each experiment (Fig.~\ref{fig:sim_results}b-c).
In addition, we ran a control experiment in which we fixed the shape of the robot to be fully inflated and oriented width-wise ($\theta=0{^\circ}$) for the inclined surface, to determine whether shape-change was necessary. %The best the robot could do was prevent itself from rolling backward, and it attained a fitness value of -0.001 BL/s. Comparing this last experiment to the others, over the inclined surface we find that given the same computational effort, the added parameters of shape and orientation increased the performance of the robot, despite the higher dimensionality of the search space.

When shape and orientation were set as fixed parameters, optimization found a rolling gait for the inflated shape on flat ground (max fitness 0.1576 BL/s; Table 1), but it was unable to discover a successful gait for the hill when in the deflated shape (max fitness -0.0023 BL/s). We were, however, able to manually program an inchworm gait (0.0246 BL/s) that enabled the deflated robot to climb the incline (Table~\ref{tbl: simulation results}). For reference, other robots that exclusively utilize inchworm gaits have widely varied speeds, ranging from 0.013 BL/s (for a 226 mm-long robot) ~\cite{felton_robot_2013}to 0.16 BL/s (for a centimeter-scale robot) ~\cite{lee_design_2011}.

For the second pair of experiments (orientation fixed), the best robots produced inflated shapes that rolled over flat ground (max fitness 0.2000 BL/s),
and similarly failed to produce inchworm motion on the inclined ground (max fitness -0.0004 BL/s).

In the last pair of experiments (all parameters open), the algorithm again discovered that cylindrical rolling robots were the most effective over a flat surface. However, over the inclined surface, the optimization algorithm found
designs with a
inflated
shape capable of shuffling up the hill when oriented at an angle (max fitness 0.0158 BL/s). Using this strategy, the robot achieved combined locomotion of 0.1023 BL/s, outperforming the hand-designed strategy of using crawling on inclines and rolling on flat ground (combined max of 0.0998 BL/s).
The robot was able to inflate two of its external bladders to balance itself like a table, increasing stability between the robot and the ground,
while the non-standard orientation reduced the amount of gravitational force opposing the direction of motion, thereby requiring less propulsive force and reducing the likelihood of the robot rolling back down the hill. However, this behavior is physically unrealistic. Thus, the best strategy for moving up the incline
remains the hand-designed flattened shape which traverses
the hill using an inchworm-like gait.
For the successful evolutionary trials,
the policies found were less finely tuned than those that were hand-designed.
Thus, even in the highly optimized rolling behaviors, there were
occasional counterproductive or superfluous actuations (see Supplementary Movie S1). Such unhelpful motions could likely be reduced via further optimization and by adding a fitness penalty for the number of actuators used per time step.

A similar trend is shown in Fig.~\ref{fig:sim_results}b-c, where the best combined robot for each environment was discovered by the pair of experiments in which the hill climbing algorithm had the most control over the optimization of the robot, despite the larger number of trainable parameters, and thus an increased likelihood of getting stuck in a local minimum. Additionally, the population of simulated robots continued to exhibit similar (and often superior) mean performance compared to the control-only experiments (Fig.~\ref{fig:sim_results}). These observations suggest that the robots avoided local minima, and that more parameters should be mutable during automated design of shape-changing robots.
We hypothesize that maximizing the algorithm's design freedom would be even more important when designing robots with increased degrees of freedom, using more sophisticated optimization algorithms that can operate in an exponentially growing search space.

Overall, this sequence of experiments showed that automated search could discover physically realistic shapes and controllers for our shape-changing robot in
one of the two environments we studied (flat ground).
Although the hand-designed controllers each performed comparably to the best discovered controllers in a single environment, by changing shape, the robot had a better combined average speed in both environments. Concretely, the best shape-controller pair found by hill climbing locomoted at a speeds of 0.1888 BL/s on flat ground and 0.0158 BL/s on incline, resulting in an average speed across the two environments of 0.1023 BL/s, compared to the average speed of -0.1995 BL/s for the round shape with a rolling gait and 0.0424 BL/s for the flat shape with an inchworm gait (Table~\ref{tbl: simulation results}).

%%%%%%%%%%%%%%% Transferring to a physical robot. %%%%%%%%%%%%%%%
\subsection{Transferring to a physical robot.}
% Introduce sim2real
Transferring simulated robots to reality introduces many challenges. For perfect transferal, the simulation and hardware need to have matching characteristics, including: material properties, friction modeling, actuation mechanisms, shape, geometric constraints, and range of motion. In practice, hardware and software limitations preclude perfect transferal, so domain knowledge must be used to achieve a compromise between competing discrepancies. Here, we sought to maximize the transferal of useful behavior, rather than strictly transferring all parameters. %, to accelerate the development of shape-changing robots.

% High-level summary of our instance
In simulation, we found that the same actuators could be used to create different locomotion gaits. When restricted to the cylindrical shape, we found that sequential inflation of the bladders could be used to induce rolling.
A hand designed policy enabled the flatter robots to move with inchworm motion.
To transfer such shape change and gaits to a physical robot, we created a robot which had an inflatable core, eight pneumatic surface-based actuators for generating motion, and variable-friction feet on each edge to selectively grip the environment (Fig.~\ref{fig: summary}). This suite of features allowed the robot to mirror the simulated robot's gaits, including rolling and inching. The ``hand-designed controllers'' from simulation were transferred to reality by sending the same command sequence from a PC to digital pressure regulators~\cite{booth_addressable_2018} that inflated the bladders, resulting in forward motion. However, it was found that different bladders expanded at different rates and had slightly different maximum inflation before failure, so in the experiments shown in this manuscript, the robots were manually teleoperated to approximate the hand-designed controllers with nonuniform timesteps between each actuation state. Further details on the robot hardware are presented in the Methods section.

\begin{figure*}
    \centering
    \includegraphics[width=0.95\textwidth]{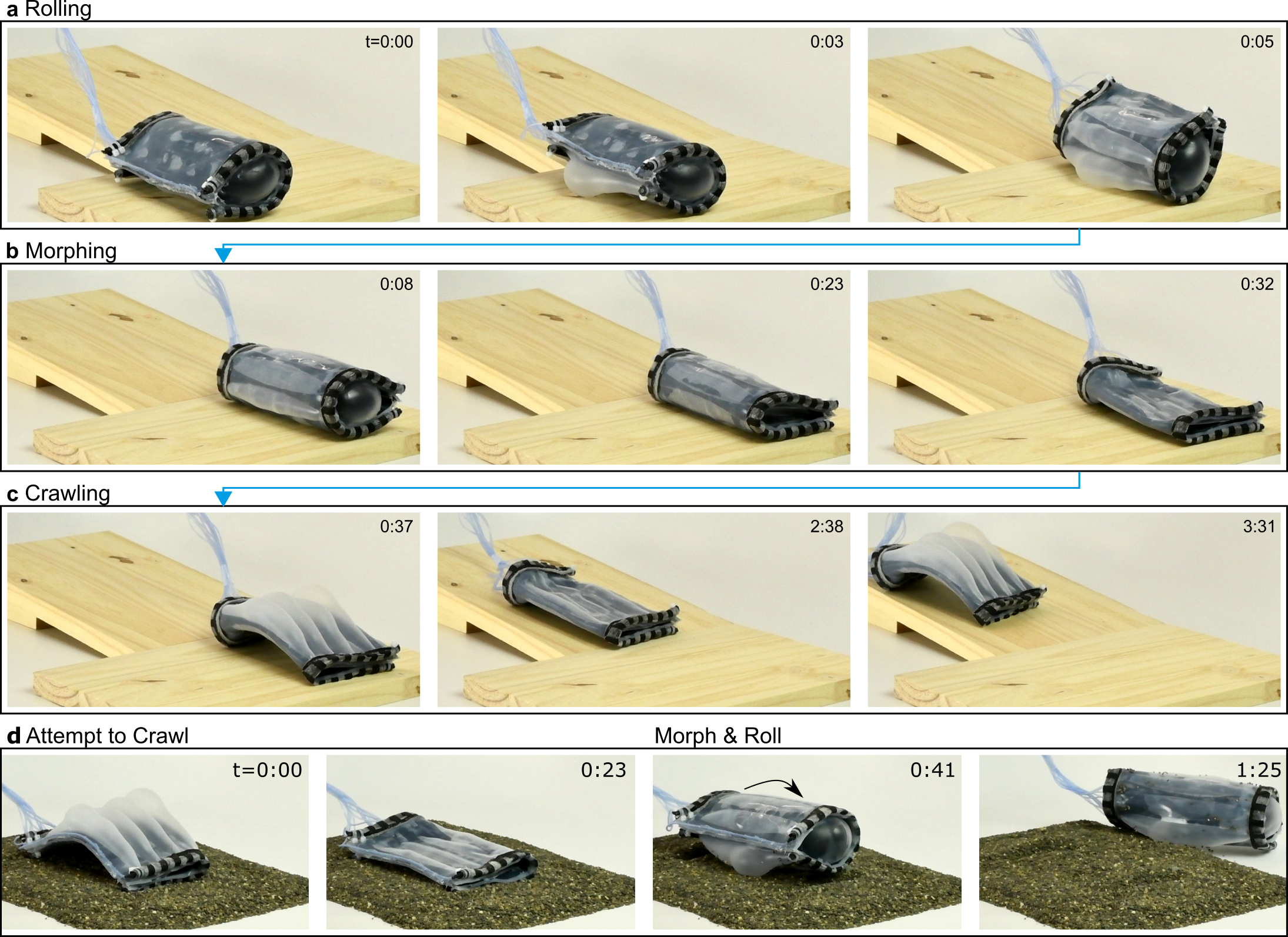}
    \caption{ \textbf{Shape change allowed the physical robot to operate in previously inaccessible environments.} \textbf{a,} While round, the robot's actuators created a rolling gait which was effective on flat ground. \textbf{b,} By deflating its inner bladder, the robot could flatten. \textbf{c,} While flat, the outer bladders induced an inchworm-like gait, allowing the robot to ascend inclines up to ${\sim}14$~degrees. \textbf{d,} The inchworm gait gripped the ground to crawl forward, making it ineffective on granular surfaces. When faced with such a situation, the robot could expand its inner bladder to begin rolling. For length-scale reference, the robot is 10 cm by 15 cm while flattened, and 7~cm diameter by 15~cm while round. Panels \textbf{a–c} correspond to times from a single trial, while panel \textbf{d} is from a different trial and has a separate start time.}
    \label{fig: real_motion}
\end{figure*}

% Rolling
Mirroring simulation, rolling was achieved by inflating the trailing edge bladder to push the robot forward, exposing new bladders that were then inflated one at a time, sequentially (Fig.~\ref{fig: summary}a and Fig.~\ref{fig: real_motion}a).  Each inflation shifted the robot's center of mass forward so the robot tipped in the desired direction, allowing the robot to roll repeatedly. This motion was effective for locomoting over flat ground (average speed 0.05 BL/s). % where a body length was the robot's longest edge length). % 0.7~cm/s
When we attempted to command the robot to roll up inclines, the slope of the incline and the robot's seam made it difficult for the robot to roll. These observations suggest the existence of a transition regime on the physical robot, where the ideal shape-locomotion pair switches from a rolling cylinder to a flat shape with inchworm gait.
However, the boundary is not cleanly defined on the physical hardware: at increasing inflation levels approaching the strain limit of the silicone, the robot could roll up increasingly steep inclines up to ${\sim}9{^\circ}$. After just a few such cycles, the bladders would irreversibly rupture, causing the robot to roll backward to the start of the incline.

% Flattening
By accessing multiple shapes and corresponding locomotion modes, shape-changing robots can potentially operate within multiple sets of environments. For example, when our robot encountered inclines, it could switch shapes (Fig.~\ref{fig: real_motion}a-c and Supplementary Movie S1). To transition to a flattened state capable of inchworm motion, the robot would deflate its inner bladder, going from a diameter of 7~cm (width-to-thickness ratio $\gamma=1$) to an outer height of ${\sim}1.2$~cm ($\gamma{\sim}8.3$) (Fig.~\ref{fig: real_motion}b). The central portions of the robot flatten to ${\sim}7$~mm, which is approximately the thickness of the robot's materials, resulting in $\gamma{\sim}14$. During controlled tests, an average flat-to-cylinder morphing operation at 50~kPa took 11.5 seconds, while flattening with a vacuum ($-80$ kpa) took 4.7 seconds (see Methods for additional details).

% Crawling
Flattening reduced the second moment of area of the robot's cross-section, allowing the bladders' inflation to bend the robot in an arc (Fig.~\ref{fig: real_motion}c). At a first approximation, body curvature is given as $\kappa = \frac{M}{EI}$, where $M$ is the externally-induced moment, $E$ is the effective modulus, and $I$ is the axial cross-section's second moment of area. Thus, flatter robots should bend to higher curvatures for a given pressure. However, even for the flattest shape, bending was insufficient to produce locomotion: on prototypes with unbiased frictional properties, bending made the robot curl and flatten in-place.

\begin{figure*}
    \centering
    \includegraphics[width=6in]{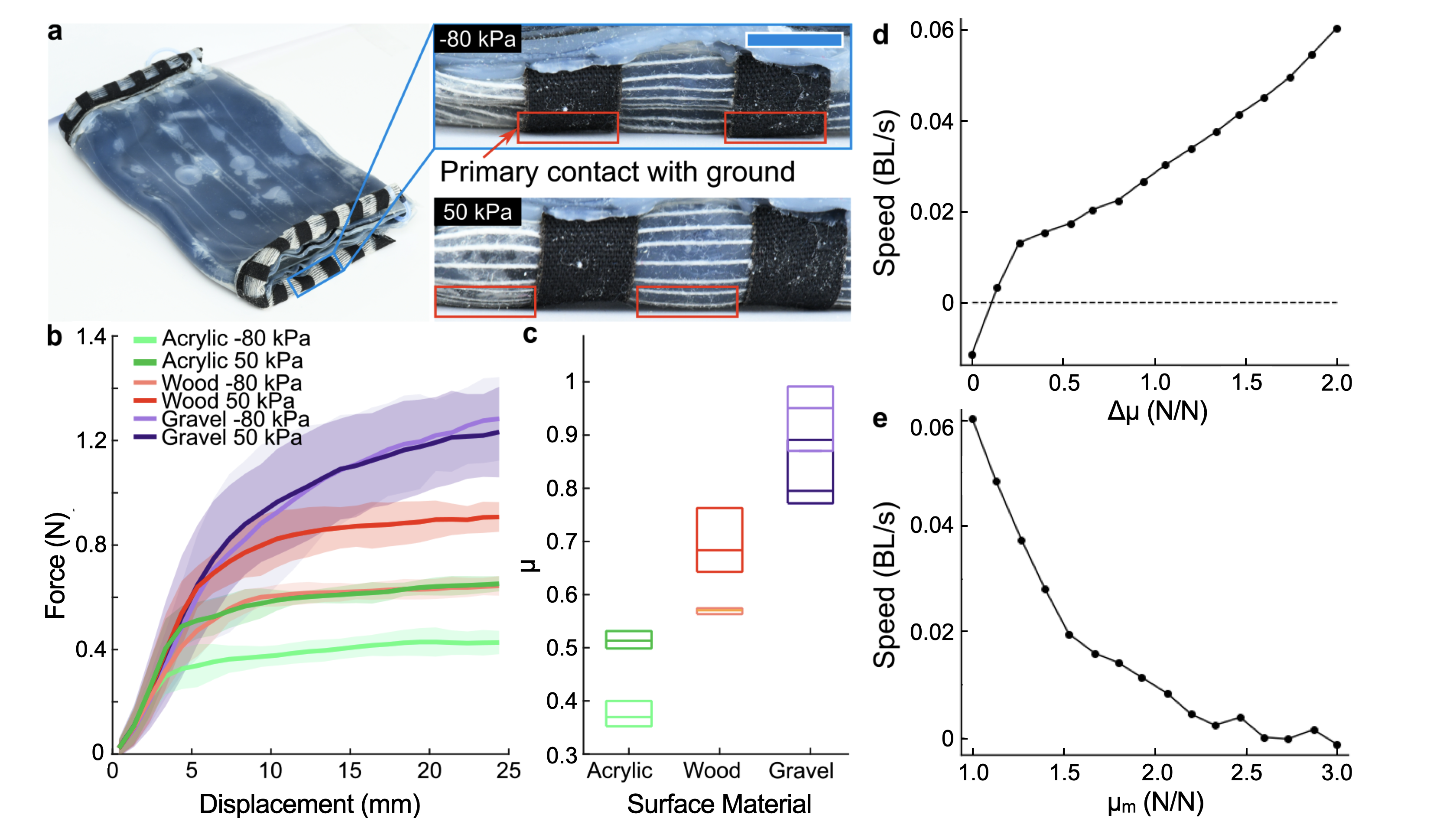}
    \caption{\textbf{The variable-friction feet change their frictional properties when inflated.} \textbf{a,} When the robot's feet are inflated, silicone bladders protrude from their fabric sheath to contact the ground. Blue scale bar on inset represents 1~cm.
    \textbf{b,} Force vs. displacement when the robot was slid over wood, acrylic, and gravel. Each shaded region represents $\pm$one standard deviation about the mean (solid line).
    \textbf{c,} Coefficient of static friction. The boxes denote 25th and 75th percentiles, while the bars represent the median.
    \textbf{d,} Speed (in simulation) as a function of the difference between friction values, $\Delta \mu = \mu_i - \mu_u$ (where $\mu_i$ is friction while the foot is inflated to 50 kPa, and $\mu_u$ is friction while uninflated at -80 kPa). \textbf{e,} Speed (in simulation) as a function of average friction value, $\mu_m = (\mu_i + \mu_u)/2$. In \textbf{d-e}, the hand-tuned inchworm gait was used.}
    \label{fig: friction}
\end{figure*}

% VF feet
Variable-friction ``feet'' were integrated onto both ends of the robot and actuated one at a time to alternate between gripping in front of the robot and at its back, allowing the robot to inch forward (average speed of 0.01~BL/s on flat wood). % 0.17~cm/s, or 
The feet consisted of a latex balloon inside unidirectionally stretchable silicone lamina~\cite{kim_reconfigurable_2019}, wrapped with cotton broadcloth. When the inner latex balloon was uninflated (-80 kPa), the silicone lamina was pulled into its fabric sheath, thus the fabric was the primary contact with the ground. When the balloon was inflated (50 kPa), it pushed the silicone lamina outward and created a higher-friction contact with the ground (Fig.~\ref{fig: friction}a). To derive coefficients of static friction ($\mu$) for both the uninflated ($\mu_u$) and the inflated ($\mu_i$) cases, we slid the robot over various surfaces including acrylic, wood, and gravel. As the robot slid over a surface, it would typically exhibit an initial linear regime corresponding to pre-slip deformation of the feet, followed by slip and a second linear kinetic friction regime (Fig.~\ref{fig: friction}b). From the pre-slip regime, we infer that on a wood surface $\mu_u$ = 0.56 and $\mu_i$ = 0.70 --- an increase of ${\sim}25\%$ (Fig.~\ref{fig: friction}c). On acrylic, $\mu_u$ = 0.38 and $\mu_i$ = 0.51, which is an increase of 35$\%$, yielding an inching speed of 0.007~BL/s. % 0.11~cm/s or 

When the difference in friction ($\Delta \mu = \mu_i - \mu_u$) for the variable-friction feet was too low (such as on gravel), inchworm motion was ineffective, as predicted by simulation (Fig.~\ref{fig: friction}d). Similarly, when the average friction ($\mu_m = (\mu_i + \mu_u)/2$) was too high, it would overpower the actuators and lead to negligible motion (Fig.~\ref{fig: friction}e). On wood, the inchworm gait was effective on inclines up to ${\sim}14{^\circ}$, at a speed of 0.008 BL/s % 0.12 cm/s or 
(Fig.~\ref{fig: real_motion} and the Supplementary Video). Thus, the robot could quickly roll over flat terrain (0.05 BL/s) then flatten to ascend moderate inclines, attaining its goal of maximizing total traveled distance. 

% Real robot can change shape to gain access to different environments

% Variable-friction feet characteristics

%%%%%%%%%%%%%%% Discussion %%%%%%%%%%%%%%%
\section{Discussion}
In this study, we tested the hypothesis that adapting the shape of a robot, as well as its control policy, can yield faster locomotion across environmental transitions than adapting only the control policy of a single-shape robot. In simulation, we found that a shape-changing robot traversed two test environments faster than an equivalent but non-morphing robot. Then, we designed a physical robot to utilize the design insights discovered through the simulation, and found that shape change was a viable and physically-realizable strategy for increasing the robot's locomotion speed. % In contrast, while inflated, the simulated robot could locomote rapidly on flat ground, and roll in the negative direction on the incline. Similarly, the inflated physical robot could roll on flat ground, but its bladders ruptured during attempts to roll up an incline. The simulated and physical robots both needed to change into a flattened shape in order to locomote up the incline.

% Sim2real
We have also shown progress toward an automated sim2real framework for realizing metamorphosing soft robots capable of operating in different environments. In such a pipeline, simulated shape-changing robots would be designed to achieve a desired function in multiple environments, then transferred to physical robots that could attain similar shapes and behaviors. We demonstrated each component of the pipeline on a representative task and set of environments: locomotion over flat ground and an incline. Starting with an initial robot design, the search method sought valid shapes and control policies which could succeed in each
environment, however, it is notable that the search found it difficult to find a successful inchworm gait. This is likely due to a lack of gradient between control policies that do not produce inching and those that do. Thus, a complete pipeline would need to include more sophisticated search methods that are better able to search this space.
The effective shapes and gaits were then transferred to physical hardware. However, the simulation was able to generate some non-transferable behavior by exploiting inaccuracies of some simulation parameters. For example,
when the friction coefficient was too low, the robot would make unrealistic motions such as sliding over the ground.
Other parameters, such as modulus, timescale, maximum inner bladder pressure, and resolution of the voxel simulation (i.e., the number of simulated voxels per bladder), and material density, could be adjusted without causing drastic changes in behavior. Developing a unified framework for predicting the sim2real transferability of multiple shapes and behaviors to a single robot remains an unsolved problem.

% real2sim
Insights from early physical prototypes were used to improve the simulator's hyperparameters (such as physical constants), resulting in more effective sim2real transferal.
Pairing hardware advances with multiple cycles through the sim2real pipeline, we plan to systematically close the loop such that data generated by the physical robot can be used to train a more accurate simulator, after which a new round of simulation to reality transfers can be attempted. This iterative process will be used to reduce the gap between simulation and reality in future experiments.

% Advances in hardware + software --> complicated shape-controller pairs
With advances such as increased control of the physical robots' shape and more efficient, parallelized soft-robot simulators, the pipeline should be able to solve increasingly challenging robot design problems and discover more complicated shape-controller pairs. While the sim2real transfer reported in this manuscript primarily tested intermediate shapes between two extremal shapes | a fully-inflated cylinder and a flattened sheet | future robots may be able to morph between shapes embedded within a richer, but perhaps less intuitive morphospace.
For example, robots could be automatically designed with a set $C$ of $N_c$ inflatable cores and corresponding constraining fabric outer layers. To transition between shapes, a different subset $C$ could be inflated, yielding $2^{N_c}$ distinct robot morphologies. Designing more sophisticated arrangements of actuators and inflatable cores could be achieved using a multilayer evolutionary algorithm, where the material properties of robots are designed along with their physical structure and control policies.~\cite{howard_evolving_2019}.
Additionally, it is unclear how to properly embed sensors into the physical robot to measure its shape, actuator state, and environment. Although some progress has been made toward intrinsically sensing the shape of soft robots~\cite{soter_bodily_2018}, and environmental sensing~\cite{umedachi_gait_2016}, it remains an open challenge for a robot to detect that it as encountered an unforeseen environment and edit its body morphology and behavioral control policy accordingly.

% Those advances --> additional environments? Generate hypotheses and SI
Future advances in hardware and search algorithms
could be used to design shape-changing robots that can operate across more challenging environmental changes.
For example, swimming or amphibious robots could be automatically designed using underwater soft-robot simulation frameworks~\cite{corucci_evolving_2018}, and changing shape within each gait cycle might allow robots to avoid obstacles~\cite{shah2019morphing} or adapt to environmental transitions. We have begun extending our framework to include underwater locomotion, where locomoting between terrestrial and aquatic environments represents a more extreme environmental transition than flat-to-inclined surface environments. 
Our preliminary results suggest that multiple swimming shape-gait pairs can be evolved using the same pipeline and robot presented herein (see Supplementary Information). 
While recent work has shown the potential advantages of adapting robot limb shape and gait for amphibious locomotion~\cite{baines_variable_2020}, closing the sim2real gap on shape-changing amphibious robots remains largely unstudied.

Collectively, this work represents a step toward the closed-loop automated design of robots that dynamically adjust their shape to expand their competencies. By leveraging soft materials, such robots potentially could metamorphose to attain multiple grasping modalities, adapt their dynamics to intelligently interact with their environment, and change gaits to continue operation in widely different environments.

%%%%%%%%%%%%%%% Methods %%%%%%%%%%%%%%%
\section{Methods}
\label{sec: Methods}

\subsection{Simulation environment.}
The robots were simulated with the multi-material soft robot simulator Voxelyze~\cite{hiller2014dynamic}, which represents robots as a collection of cubic elements called voxels. A robot can be made to move via external forces or through expansion of a voxel along one or more of its 3 dimensions.

% \begin{figure}[!ht] % Simulation details
%     \centering
%     \includegraphics[width=\textwidth]{Figures/simulation_details.png}
%     \caption{\textbf{a,} Visual rendering of the robot in both its inflated and flat shapes, with the corresponding underlying Euler-Bernoulli beams shown on the right, in red. \textbf{b,} A voxel is represented as a point mass connected to its neighbors by beams, adapted from~\cite{hiller2014dynamic}. \textbf{c,} The pressure ($P$) vectors (red) acting on each interior voxel (green) when fully inflated.}
%     \label{fig: sim_details}
% \end{figure}

Voxels were instantiated as a lattice of Euler-Bernoulli beams (Fig.S~\ref{fig: sim_details}a). Thus, adjacent voxels were represented as points connected by beams (Fig.S~\ref{fig: sim_details}b). Each beam had length $l=0.01~m$, elastic modulus $E=400~kPa$, density $\rho=3000~\frac{Kg}{m^3}$, coefficient of friction
$\mu=2.0$
, and damping coefficient $\zeta=1.0$ (critically damped). For comparison, silicone typically has a modulus of ${\sim}100-600~kPa$ and density of ${\sim}1000~\frac{Kg}{m^3}$. These parameters were initially set to $E=100~kPa$ and $\rho=1000~\frac{Kg}{m^3}$, but were iteratively changed to increase the speed and stability of the simulation while maintaining physically realistic behavior. % Dragon Skin 10 E = 151 kPa, DS30 E = 592 kPa. DS 10 & 30 density 1.06 g/cc  = 1070 kg/m^3.
% Source: https://www.smooth-on.com/tb/files/DRAGON_SKIN_SERIES_TB.pdf
We simulated gravity as an external acceleration ($g=9.80665~\frac{m}{s^2}$) acting on each voxel. For the flat environment, gravity was in the simulation's negative z direction. Since changing the direction of gravity is physically equivalent to and computationally simpler than rotating the floor plane, we simulated the slope by changing the direction of gravity.
The robot could change shape by varying the force pushing outward, along to the interior voxels' surface normals, representing a discrete approximation of pressure (Fig.S~\ref{fig: sim_details}c). % The core's inflation force ranged from 0 to $\approx1.4~N$ per voxel, representing pressure of $14 ~kPa$. 
The maximum pressure was set at
$12~kPa$
after comparison to prior results (for example, the robotic skins introduced by Shah et al. inflated their pneumatic bladders to under 20 kPa~\cite{shah2019morphing}) and after initial experiments with hardware that suggested only 10${\sim}$35 kPa was necessary. % By comparison, the bladders of the rolling soft robot in our previous work used inflation pressures of 3 psi~\cite{shah2019morphing}. % PancakeBot used 7.5 psi = 50 kPa

The robots' external bladders were simulated via voxel expansion such that a voxel expanded along the z-dimension of its local coordinate space at $3e^{-4}~m$ per simulation step and $1.5e^{-5}~m$ along the x-dimension. Expansion in the y-dimension created a bending force on the underlying skin voxels. This value was changed on a sliding scale from $1.76e^{-4}~m$ to $3e^{-5}~m$ based on the pressure of the robots' core, such that bladder expansion created minimal bending force when the robot was inflated, simulating the expansion of physically-realizable soft robots. Concretely, the y-dimension expansion was computing using a normalizing equation $(b - a) * ((P - P_{MIN})/(P_{MAX} - P_{MIN})) + a$ where $a = 1.7$, $b=10$, $P_{MAX}$ is the maximum outward force per voxel in the robot's core ($1.4~N$), $P_{MIN}$ is the minimum outward force per voxel $0~N$ and $P$ is the current outward force per voxel. These values were adjusted iteratively, until simulated and physical robots with the same controllers exhibited similar behavior in both the inclined and flat environments. Lastly, to prevent the robot from slipping down the hill, and to enable other non-rolling gaits, the robot was allowed to change the static and kinetic friction of its outer voxels between a low value ($\mu=1e{-4}$) when inactive and high value ($\mu=2.0$) when active.

\subsection{Optimization}
The optimization algorithm searched over 3 adjustable aspects of the robots: shape (parameterized as inner bladder pressure), orientation of the robot relative to the incline, and actuation sequence. The algorithm searched over a single number
$p\in[0, 12]$ ($kPa$)
for shape and $\theta\in[0^{\circ}, 90^{\circ}]$ for orientation (see Fig.S~\ref{fig:sim_results}a for illustrations of each parameter). % $0^{\circ}$ faces the robot up the incline, while $90^{\circ}$ places it sideways on the slope

The robot's actuation sequence $S$ over $T$ actuation steps was represented by a binary $10 \times T$ matrix where a $1$ corresponds to bladder expansion and $0$ corresponds to bladder deflation. Each of the first eight rows corresponded to one of the inflatable bladders, and the last two rows controlled the variable friction feet. Each column represented the actuation to occur during a discrete amount of simulation time steps $t$, resulting in a total simulation length of $t*T$. $t$ was set such that an actuation achieved full inflation, followed by a pause for the elastic material to settle. Actuating in this manner minimizes many effects of the complex dynamics of soft materials, reducing the likelihood of the robots exploiting idiosyncraies of the simulation environment.
In this study, we used $t=11670$ timesteps of $\approx$0.000106 ($dt$) seconds each and $T = 8$ for all simulations. We the actuation matrix $T$ is repeated twice for each simulation thus doubling the length of simulation for a total simulation time of $\approx$19.79 seconds ($(t*dt)*2T)$.
To populate $S$, the algorithm searched over a set of parameters (frequency $f$ and offset $\phi$) for each of the ten actuators. Both of these parameters were kept in the range $0-T$ where in our case we set $T=16$. $f$ determined the number of columns between successive actuator activations, where $f=0$ created a row in the actuation matrix of all $1$'s, $f=1$ created a row with every other column filled by a $1$, $f=2$ every two columns filled by a $1$, and so on. $\phi$ specified the number of columns before that actuator's first activation.

We optimized the parameters of shape, orientation and actuation using a hill climber method. This method was chosen for computational efficiency, since a single robot simulation took considerable wall-clock time (approximately 2.5 minutes on a 2.9 GHz Intel Core i7 processor). The hill-climber algorithm needs only one robot evaluation per optimization step, in contrast to more advanced optimization algorithms that often require multiple evaluations per optimization step. The current set of parameters $C$ was initialized to randomly-generated values and evaluated in the simulation, where fitness was defined as the distance traveled over flat ground, or distance traveled up the incline. A variant $V$ was made by mutating each of the parameters by sampling from a normal distribution centered around the current parameters of $C$.
$V$ was then tested in the simulation, and if it traveled farther, the algorithm replaced $C$ with $V$ and generated a new $V$. The process of generating variations, evaluating fitness, and replacing the parameters was done for 200 generations.
To determine the repeatability of such an algorithm, we ran 60 independent hill climbers for each of the 6 experiments, as described in the Results section.

%%%% Physical Robot
\subsection{Manufacturing the physical robot.}
The physical robot was designed to enable transfer of function, shapes, and control policies from simulation, while maximizing locomotion speed and ease of manufacture. In summary, the inner bladder was silicone (Dragon Skin 10, abbreviated here as DS10, Smooth-On Inc.), the cylindrical body was cotton dropcloth, and the outer bladders were made with a stiffer silicone (Dragon Skin 30, abbreviated here as DS30, Smooth-On Inc.) for higher force output. The variable-friction feet were made out of latex balloons, unidirectionally stretchable lamina (STAUD prepreg, described in~\cite{kim_reconfigurable_2019}), and cotton dropcloth. Complete manufacturing details follow.

First, the outer bladders were made (Fig.S~\ref{fig:manufacturing}a). Two layers of DS10 were rod coated onto a piece of polyethylene terephthalate (PET). After curing, the substrate was placed in a laser cutter (ULS 2.0), PET-side up, and an outline of the eight bladders were cut into the PET layer. The substrate was removed from the laser cutter and the PET not corresponding to the bladders (i.e., the outer ``negative'' region) was removed. Two layers of DS30 were rod coated onto the substrate. DS30 is stiffer than DS10, and was used to increase the outer actuators' bending force, while DS10 was used in all other layers to keep the robot flexible. Using ethanol as a loosening agent, the encased PET was then removed from all eight bladders. Finally, a layer of DS10 was cast over the bladders' DS10 side for attaching broadcloth to begin manufacturing of the inner bladder.

% % Manufacturing the real robot
% \begin{figure*}
%     \centering
%     \includegraphics[width=0.95\textwidth]{Figures/manufacturing.png}
%     \caption{ \textbf{Manufacturing the physical robot.} 
%     \textbf{a,} First, the outer bladders were made out of silicone. \textbf{b,} The outer bladders were bonded to silicone-soaked cotton broadcloth, and the inner bladder was fabricated. \textbf{c,} To make variable-friction feet, rectangular slits were lasercut into broadcloth, and unidirectionally stretchable lamina~\cite{kim_reconfigurable_2019} and latex baloons were attached with silicone. \textbf{d,} The robot was assembled by attaching the feet to the main robot body, and the robot was folded to bond the inner bladder to the bladder-less half.
%     Small squares to the right of the schematic for each step depicts a simplified cross-sectional view of the robot at the end of that step}.
%     }
%     \label{fig:manufacturing}
% \end{figure*}

The inner bladder was made by first soaking cotton broadcloth (15 cm by 20 cm) with DS10, and placing it on the uncured layer on top of the outer bladders (Fig.S~\ref{fig:manufacturing}b). PET was then laid on the robot, and the inner bladder outline was lasercut into the PET. Again, the outer PET was removed, and DS10 was rodcoated to complete the inner bladder. The PET was removed using ethanol and tweezers, and silicone tubing (McMaster-Carr) was inserted into each bladder and adhered with DS10.

To make the variable-friction feet, rectangular slits were lasercut into broadcloth, and unidirectionally stretchable laminate~\cite{kim_reconfigurable_2019} was attached using Sil-Poxy (Smooth-On Inc.) (Fig.S~\ref{fig:manufacturing}c). Latex balloons were attached using Sil-Poxy, and the feet were sealed in half with Sil-Poxy to make an enclosed envelope for each foot. When at vacuum or atmospheric pressure, the fabric would contact the environment, leading to a low-friction interaction. When the feet were inflated, the silicone would contact the environment, allowing the feet to increase their friction.

Finally, the robot was assembled by attaching the feet to the main robot body using Sil-Poxy, and the robot was folded to bond the inner bladder to the bladder-less half, using DS10 (Fig.S~\ref{fig:manufacturing}d).

\subsection{Experiments with the physical robot.}
To test the robot's locomotion capabilities, we ran the physical robots through several tests on flat and inclined ground. The pressure in the robots' bladders was controlled using pneumatic pressure regulators~\cite{booth_addressable_2018}. The robots were primarily operated on wood (flat and tipped to angles up to ${\sim}15^\circ$), with additional experiments carried out on a flat acrylic surface and a flat gravel surface (see Fig.~\ref{fig: real_motion} and the Supplementary Movie S1).

The variable-friction feet were assessed by pulling the robot across three materials (acrylic, wood, gravel) using a materials testing machine (Instron 3343). The robot was placed on a candidate material and dragged across the surface at $100~mm/min$ for $130~mm$ at atmospheric conditions ($23^\circ~ C$, $1~atm$). This process was repeated 10 times for each material, at two feet inflation pressures: vacuum ($-80~kPa$) and inflated ($50~kPa$). The static coefficient of friction, $\mu_s$, was calculated by dividing the force at the upper end of the linear regime by the weight of the robot.

The robot's shape-changing speed was assessed by manually inflating and deflating the robot's inner core for 20 cycles. For each cycle, the robot body was inflated to a cylindrical shape with a line pressure of $50~kPa$, and the time required to attain a diameter of ${\sim}7~cm$ was recorded. The body was then deflated with a line pressure of $-80~kPa$, and the time required to flatten to a height of ${\sim}1.2~cm$ was recorded. % 82.5 kPa

%%%%%%%%%%%%%%% Other misc. sections %%%%%%%%%%%%%%%
\section*{Corresponding Author}
% Please include a statement before the Acknowledgements naming the **author to whom correspondence** and requests for materials should be addressed.
\noindent  Correspondence and requests for materials should be addressed to R.K.B.

\section*{Acknowledgments}
This work was supported by NSF EFRI award 1830870.
Dylan Shah was supported by a NASA Space Technology Research Fellowship (80NSSC17K0164).
Joshua Powers was supported by the Vermont Space Grant Consortium under NASA Cooperative Agreement NNX15AP86H.

\section*{Author contributions}
J.B., R.K.B., S.K., D.S. and J.P. conceived the project and planned the experiments. 
J.P. coded the simulation and ran the evolutionary algorithm experiments. 
D.S. and L.T. manufactured the robot and performed the hardware experiments.
D.S., J.P., L.T., S.K., J.B., and R.K.B. drafted and edited the manuscript.
All authors contributed to, and agree with, the content of the final version of the manuscript.
% Correspondence and requests for materials should be addressed to R.K.B. and J.B..

\section*{Data Availability}
The data that support the findings of this study are available from the corresponding author upon reasonable request.

\section*{Code Availability}
A public repository at
\hyperlink{https://github.com/jpp46/NATURE_MI2020}{github.com/jpp46/NATURE\_MI2020}
contains the code necessary to reproduce the soft-robot simulations.

\section*{Competing Interests}
The authors declare no competing interests.

%%%%%%%%%%%%%%% References %%%%%%%%%%%%%%%
\bibliographystyle{naturemag}
\bibliography{references.bib}
% \printbibliography

%%%%%%%%%%%%%%% Supplementary Materials %%%%%%%%%%%%%%%

% Prepare for Supplementary Materials
\FloatBarrier % Prevent main-text figures from spilling over
\newpage
\renewcommand{\thepage}{S\arabic{page}} 
\renewcommand{\thesection}{S\arabic{section}}  
\renewcommand{\thetable}{S\arabic{table}}  
\renewcommand{\thefigure}{S\arabic{figure}}
\setcounter{figure}{0}
\setcounter{section}{0}
\setcounter{page}{1}

%%%%%%%%%% Supplementary Materials %%%%%%%%%%
\section*{Supplementary Material}
\label{sec: Supplementary Material}

\begin{itemize} %[noitemsep,topsep=0pt,parsep=0pt,partopsep=0pt]
    \item Movie S1. A soft robot that adapts to environments through shape change
    \item Figure S1. The simulated robot
    \item Figure S2. Manufacturing the physical robot
\end{itemize}

\begin{figure}[!ht] % Simulation details
    \centering
    \includegraphics[width=\textwidth]{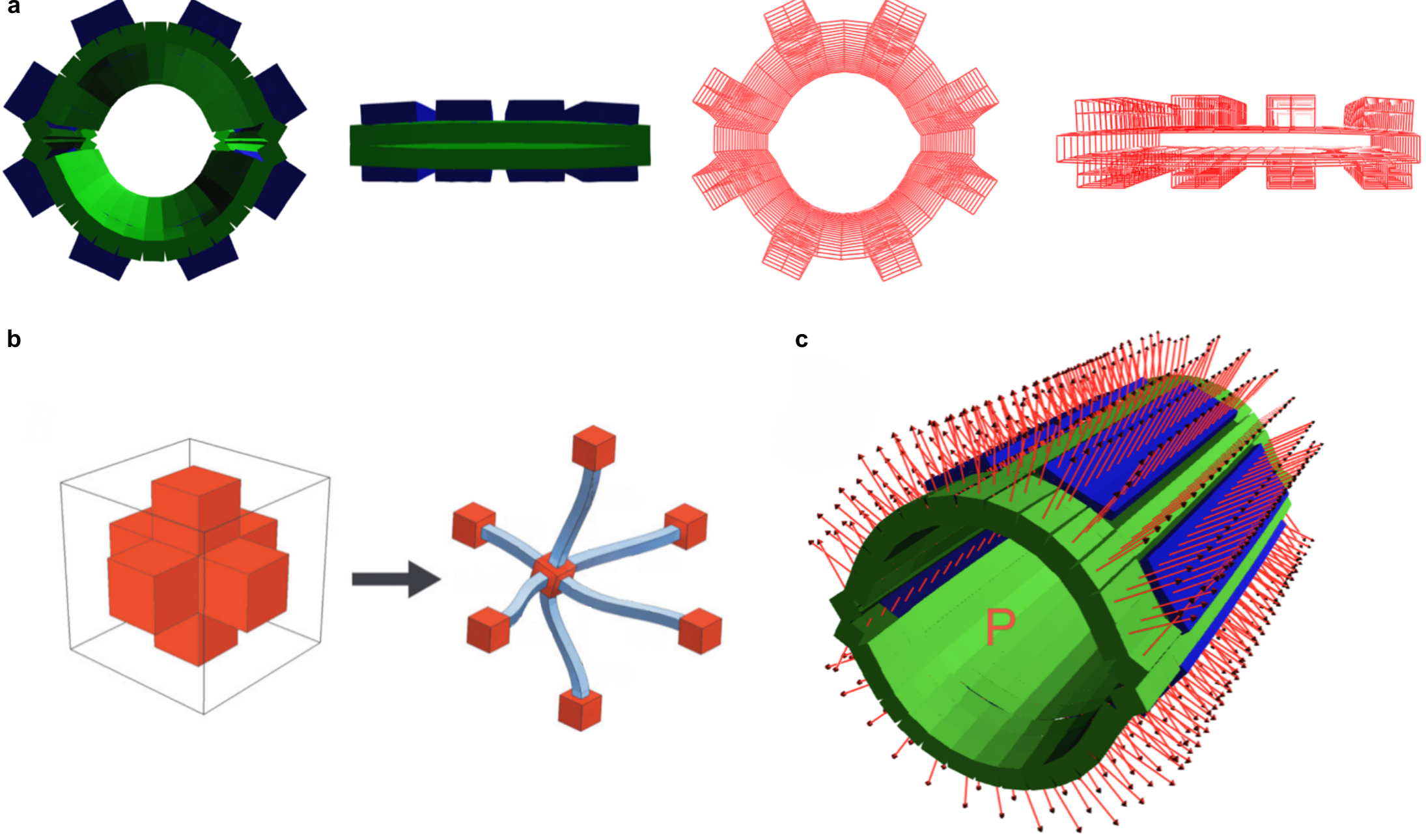}
    \caption{\textbf{The simulated robot could switch between round and flat shapes, as modeled by a shape-changing lattice.} \textbf{a,} Visual rendering of the robot in both its inflated and flat shapes, with the corresponding underlying Euler-Bernoulli beams shown on the right, in red. \textbf{b,} A voxel is represented as a point mass connected to its neighbors by beams, adapted from~\cite{hiller2014dynamic}. \textbf{c,} The pressure ($P$) vectors (red) acting on each interior voxel (green) when fully inflated.}
    \label{fig: sim_details}
\end{figure}

% Manufacturing the real robot
\begin{figure*}
    \centering
    \includegraphics[width=0.95\textwidth]{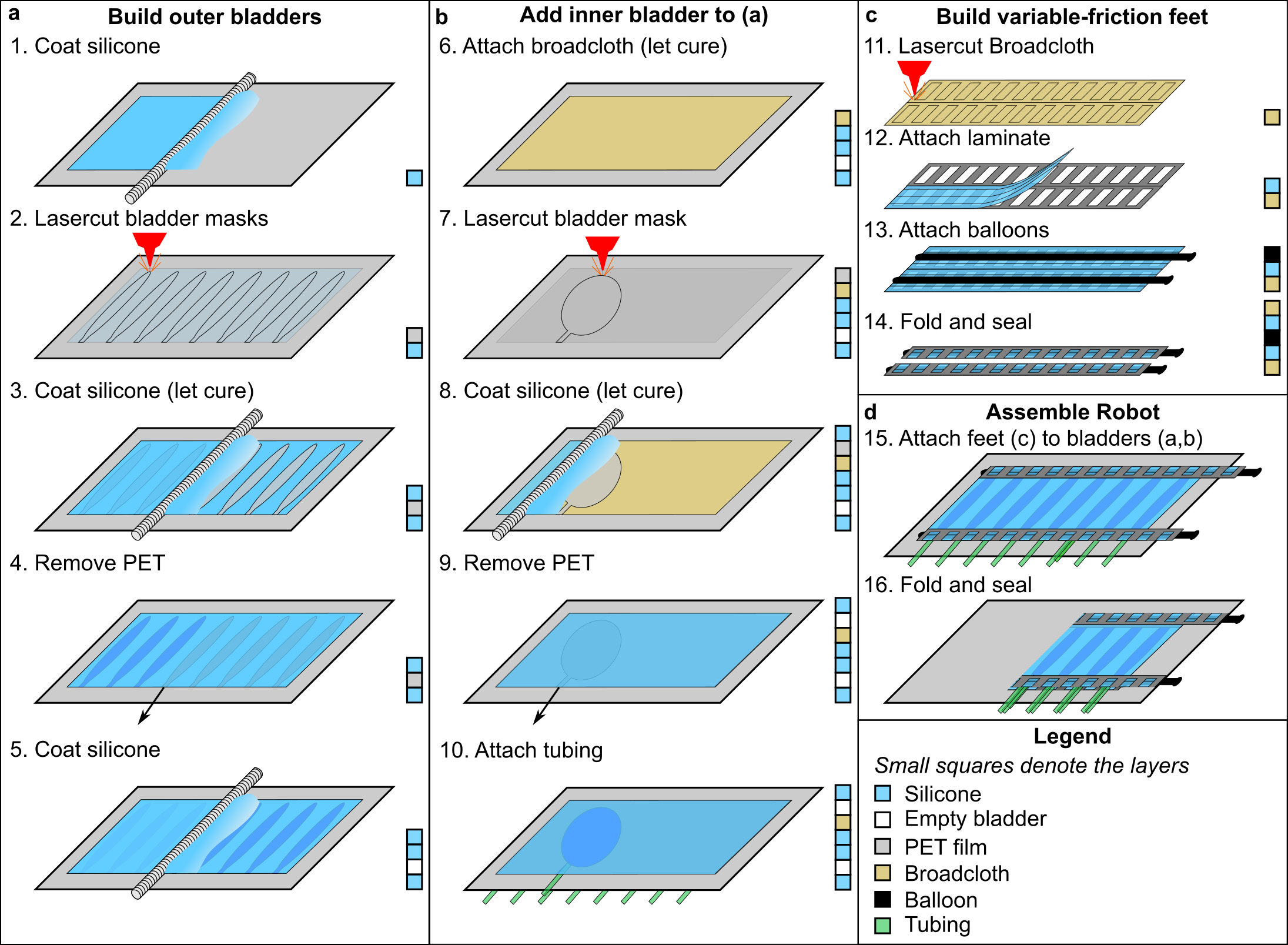}
    \caption{ \textbf{Manufacturing the physical robot.} 
    \textbf{a,} First, the outer bladders were made out of silicone. \textbf{b,} The outer bladders were bonded to silicone-soaked cotton broadcloth, and the inner bladder was fabricated. \textbf{c,} To make variable-friction feet, rectangular slits were lasercut into broadcloth, and unidirectionally stretchable lamina~\cite{kim_reconfigurable_2019} and latex balloons were attached with silicone. \textbf{d,} The robot was assembled by attaching the feet to the main robot body, and the robot was folded to bond the inner bladder to the bladder-less half.
    Small squares to the right of each schematic depict a simplified cross-section of the robot.
    }
    \label{fig:manufacturing}
\end{figure*}

%%%%%%%%%%%%%%% Response to reviewers %%%%%%%%%%%%%%%
% \input{First_revision/response_to_reviewers.tex} % Response to reviewers
% \input{Second_revision/response_to_reviewers.tex} % Response to reviewers

\end{document}